\documentclass[11pt]{article}

\makeatletter
\let\showhyphens\@undefined
\makeatother

\usepackage[final]{acl}
\usepackage{natbib}  % ← 加这行
\usepackage{amssymb}
\usepackage{hyperref}
\usepackage{microtype}
\usepackage{float}
\usepackage[ruled,vlined,linesnumbered]{algorithm2e}

\usepackage[english,bidi=default]{babel}
\babelfont{rm}{TeXGyreTermesX}
\babelprovide[import]{hindi}
\babelfont[*devanagari]{rm}{Lohit Devanagari}
\babelprovide[import]{arabic}
\babelfont[*arabic]{rm}{Noto Sans Arabic}

\usepackage{booktabs}
\usepackage{multirow}
\usepackage{amsmath}
\usepackage{graphicx}
\usepackage{subcaption}  

\title{
SAID: Accelerating Diffusion-Based Language Models \\
via Scaffold-Aware Iterative Decoding
}

\author{
Na Li$^{*1}$,
Chengda Wang$^{*1}$,
Mingju Gao$^{1}$,
Hao Tang$^{\dagger1}$ \\
\\[-0.3em]
$^{1}$School of Computer Science, Peking University \\
\\[-0.3em]
\texttt{natalie60209@gmail.com, haotang@pku.edu.cn}
}

\begin{document}

\maketitle

\begingroup
\renewcommand\thefootnote{}
\footnotetext{
$^*$Equal contribution. 
$^\dagger$Corresponding author.
}
\endgroup

\begin{abstract}
Diffusion large language models (DLLMs) enable non-autoregressive generation by iteratively denoising corrupted token sequences with bidirectional context. Despite their ability to update multiple positions in parallel, inference remains costly due to the many denoising steps required for high-quality generation. We propose \textbf{SAID}, a \underline{S}caffold-\underline{A}ware \underline{I}terative \underline{D}ecoding framework that accelerates DLLMs by reallocating computation across tokens. SAID first spends denoising computation on scaffold tokens to establish the coarse semantic structure, and then completes predictable detail tokens with fewer steps. We further adapt SAID to block-wise diffusion decoding and introduce Confidence-Hierarchical Layered Generation (CHLG), which assigns additional steps only to low-confidence tokens. Experiments on LLaDA-8B and LLaDA 1.5 across math, coding, and knowledge benchmarks show that SAID significantly accelerates DLLM inference with a maximum speedup of 9.1× while maintaining competitive performance. Our code is publicly available: \href{https://github.com/TH-AI-Lab-PKU/SAID}{https://github.com/TH-AI-Lab-PKU/SAID}.
\end{abstract}

%-----------------------------------------------------------------------
\section{Introduction}
%-----------------------------------------------------------------------

% Autoregressive (AR) language models such as GPT and LLaMA generate text by predicting tokens sequentially from left to right. Although highly effective, this sequential generation scheme fundamentally limits parallelism during inference and makes it difficult to revise earlier decisions in light of later context. Recent work on speculative decoding and parallel decoding attempts to mitigate latency, but the inherent sequential dependency chain remains a bottleneck~\citep{leviathan2023fast,stern2018blockwise}.

% Diffusion language models, originally motivated by continuous-domain diffusion
% processes~\citep{ho2020ddpm}, have been adapted to discrete text generation~\citep{austin2021d3pm,gong2022diffuseq,nie2025llada}. Models such as MDLM and LLaDA applies forward--backward diffusion processes over token sequences, progressively denoising a corrupted sequence into coherent text. The bidirectional attention used by these models enables globally coherent generation and flexible conditioning. However, the large number of denoising steps required at inference time significantly increases latency relative to single-pass AR decoding.

\begin{figure*}[ht]
  \centering
  \includegraphics[width=\linewidth]{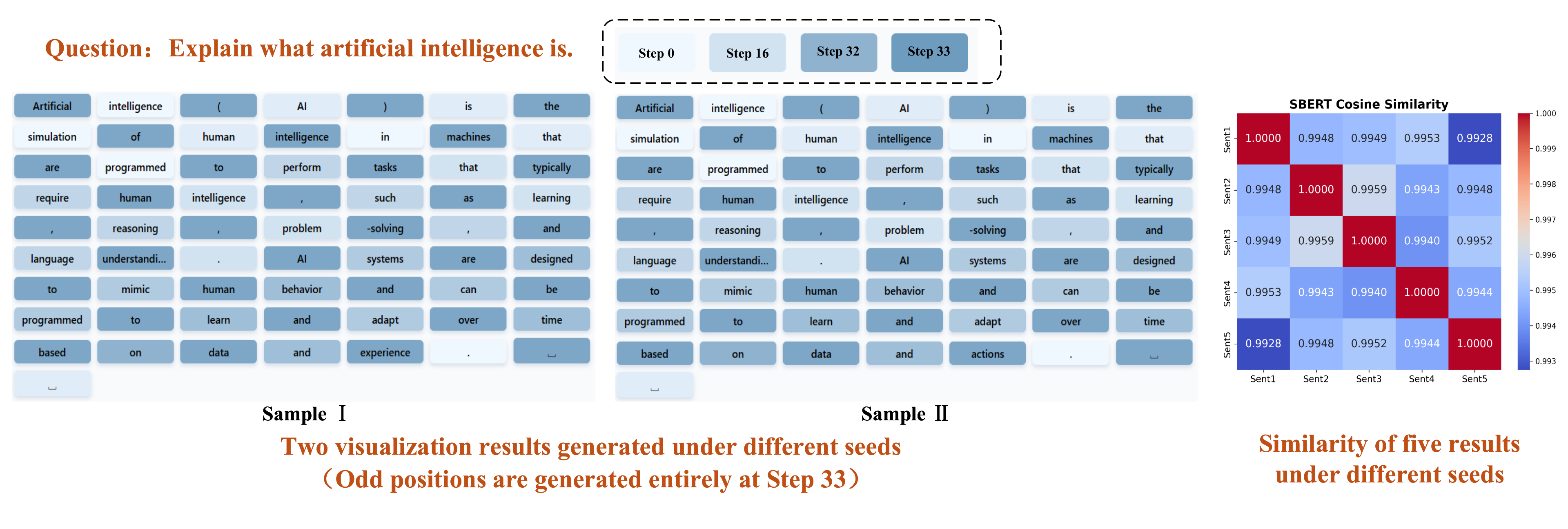}
  \caption{After determining even positions, set Temperature to 1.0 to perform five decoding runs for odd positions, and calculate their cosine similarity.}
  \label{fig:motivation}
  \vspace{-10pt}
  
\end{figure*}

Large language models (LLMs) have become the dominant paradigm for text generation~\citep{achiam2023gpt, touvron2023llama, touvron2023llama2}. 
As these models are increasingly deployed in interactive applications, inference latency has become a key challenge~\citep{zhou2024survey}. 
Most widely used models, such as GPT~\cite{achiam2023gpt} and LLaMA~\cite{touvron2023llama, touvron2023llama2}, generate text autoregressively by predicting tokens from left to right. While effective, this decoding paradigm imposes a sequential dependency chain that limits parallelism and makes it difficult to revise earlier predictions using later context. Existing acceleration methods, including speculative decoding~\cite{leviathan2023fast, chen2023speculative} and parallel decoding~\cite{stern2018blockwise, arriola2025bd3}, alleviate latency to some extent, but they remain constrained by the left-to-right generation process~\citep{leviathan2023fast,stern2018blockwise}.

These limitations have motivated non-autoregressive generation paradigms that better exploit parallel computation. Diffusion language models (DLLMs), inspired by continuous-domain diffusion processes~\citep{ho2020ddpm}, generate text by iteratively denoising a corrupted sequence~\citep{austin2021d3pm,gong2022diffuseq,nie2025llada, ye2025dream}. Unlike AR models that irrevocably fix tokens one by one, DLLMs can update multiple positions in parallel and refine earlier predictions across denoising steps using bidirectional context. Although this parallel refinement can improve generation efficiency, DLLMs still require many denoising steps at inference time, which remains a major obstacle to low-latency generation.

\textbf{Not All Tokens Need Equal Computation.}
To reduce the inference cost of DLLMs, we revisit how computation is allocated during iterative decoding. Existing decoding procedures often process unresolved positions in a largely \textbf{uniform} manner, implicitly assigning similar denoising effort to all tokens. However, we find that the value of computation varies significantly across positions. Once a coarse semantic scaffold is established, many remaining tokens become strongly constrained by the existing context and can be resolved with much less computation. As shown in Figure~\ref{fig:motivation}, when a subset of scaffold tokens is fixed, e.g., the even-position tokens in our teaser example, the remaining tokens are largely stable across different random seeds. This motivates a scaffold-aware decoding strategy: spend more computation on establishing the semantic structure early, and use fewer denoising steps for predictable detail tokens.

% Several methods have been proposed to accelerate diffusion inference, including block-partitioning strategies~\citep{arriola2025bd3,wu2025fastdllm} and improved unmasking schedules. The GTR framework was originally proposed for visual masked autoregressive (MAR) models~\citep{li2024gtr}, achieving a 3.72$\times$ speedup on MAR-H (ImageNet class-conditional generation) with negligible quality loss. GTR decomposes generation into a slow, conservative \emph{generation stage} that establishes global semantic structure, and a fast \emph{reconstruction stage} that completes remaining tokens in very few steps. The checkerboard-style stage partitioning ensures that each ungenerated token is surrounded by already-generated neighbors, strongly constraining its distribution and enabling highly parallel reconstruction.

Motivated by this observation, we propose \textbf{SAID}, a \underline{S}caffold-\underline{A}ware \underline{I}terative \underline{D}ecoding framework for accelerating diffusion-based language models. Instead of allocating denoising computation uniformly to all positions, SAID first applies the standard DLLM iterative decoding process to a subset of scaffold tokens, to establish the coarse semantic structure of the sequence. Conditioned on this fixed scaffold, the remaining detail tokens are then generated with substantially fewer denoising steps. To make SAID compatible with recent block-wise DLLM inference paradigms, we further adapt it to block diffusion decoding by applying scaffold construction and detail generation within each block. Finally, since some detail tokens may still remain uncertain after this efficient completion stage, we introduce \textbf{Confidence-Hierarchical Layered Generation} (CHLG), which assigns additional denoising steps to low-confidence tokens while committing high-confidence tokens early.

In our experiments, we evaluate SAID on LLaDA-8B and LLaDA 1.5 across math, coding, and knowledge benchmarks~\citep{cobbe2021gsm8k,
austin2021mbpp, rein2023gpqa, clark2018arc, hendrycks2021math, wang2024mmlupro}. Results show that SAID substantially accelerates DLLM inference while maintaining comparable generation quality, demonstrating its effectiveness on both standard and block-wise diffusion decoding settings. Our contributions can be summarized as follows:
\begin{itemize}
    \item We propose \textbf{SAID}, a scaffold-aware iterative decoding framework that accelerates diffusion-based language models by allocating more denoising computation to semantic scaffold construction and fewer steps to predictable detail tokens.

    \item We adapt SAID to block-wise diffusion decoding by applying scaffold construction and detail generation within each block, making it compatible with recent block diffusion language models.

    \item We introduce \textbf{Confidence-Hierarchical Layered Generation} (CHLG), which assigns additional denoising steps to low-confidence tokens while committing high-confidence tokens early.

    \item We conduct experiments on LLaDA-8B and LLaDA 1.5 across math, coding, and knowledge benchmarks, showing that SAID achieves substantial inference speedups while maintaining comparable generation quality.
\end{itemize}

%-----------------------------------------------------------------------

%-----------------------------------------------------------------------
\section{Related Work}
%-----------------------------------------------------------------------
 
\subsection{Diffusion Language Models}
 Autoregressive language models generate text token by token, a paradigm that
is inherently sequential and prevents revision of earlier decisions in light of
later context. Diffusion models, which iteratively refine noisy observations into
coherent outputs~\citep{ho2020ddpm,song2021score}, offer a structurally different
alternative: bidirectional attention over the full sequence allows each position
to condition on global context, and the iterative refinement process naturally
supports revision across steps. Adapting this framework to discrete tokens,
however, requires rethinking the corruption process. Continuous diffusion relies
on Gaussian perturbations that do not transfer to symbolic data;
\citet{sohl2015deep} and \citet{austin2021d3pm} addressed this by formulating
the forward process as a discrete-state Markov chain---the D3PM framework---and
\citet{campbell2022ctmc} later extended it to continuous-time transitions for
greater flexibility.

A particularly effective instantiation of D3PM is the masked diffusion paradigm,
which corrupts sequences by replacing tokens with a \texttt{[MASK]} symbol and
trains a bidirectional model to reconstruct the originals. The simplicity of
this absorbing-state process belies its empirical strength: SEDD~\citep{lou2024sedd},
MDLM~\citep{sahoo2024mdlm}, and RADD~\citep{ou2025radd} established that
compact masked diffusion models are competitive language models, and subsequent
scaling has widened their applicability. LLaDA~\citep{nie2025llada} demonstrates
that masked diffusion at the 8B parameter scale matches autoregressive baselines
on reasoning and code generation, while Dream~\citep{ye2025dream} and
MMaDA~\citep{yang2025mmada} push the paradigm further into improved noise schedules
and multimodal generation respectively.

\subsection{Acceleration of Diffusion Language Models}

Reducing the inference latency of DLLMs is critical for practical deployment,
since the iterative denoising process incurs far more computation than a single
autoregressive forward pass. Existing acceleration strategies fall into two
complementary lines of work: optimizing how key-value states are cached across
denoising steps, and redesigning the token unmasking procedure itself.

\noindent\textbf{Dynamic Caching Mechanisms.}
Caching in diffusion models is fundamentally more challenging than in autoregressive
models~\citep{pope2023efficiently}, because bidirectional attention means that KV
representations change whenever any token is updated. Fast-dLLM~\citep{wu2025fastdllm}
addresses this through chunked KV caching combined with confidence-aware parallel
decoding, allowing approximate reuse across steps without sacrificing generation
quality. BD3-LMs~\citep{arriola2025bd3} take a different approach by imposing a
strict left-to-right block ordering, under which all preceding blocks are fully
committed before the next begins, enabling exact KV reuse at the cost of some
generation flexibility. dKV-Cache~\citep{ma2025dkvcache} proposes a delayed caching
strategy that exploits the observation that different tokens exhibit distinct KV
representation dynamics across denoising steps, deferring cache updates until
predictions stabilize and achieving 2--10$\times$ inference speedup. d$^2$Cache~\citep{jiang2025d2cache}
further introduces a dual adaptive caching framework that fine-grainedly selects
which token KV states to update at each step, achieving speedups while also
improving generation quality.

\noindent\textbf{Optimized Sampling Strategies.}
A parallel line of work targets the unmasking schedule directly.
Confidence-based decoding~\citep{chang2022maskgit} commits tokens in decreasing
order of model certainty, reducing error propagation relative to random unmasking.
WINO~\citep{hong2025wino} introduces a revocable decoding algorithm that
aggressively drafts multiple tokens in parallel and re-masks low-confidence
positions using bidirectional context for verification, achieving up to $6\times$
speedup on GSM8K with improved accuracy. COVER~\citep{cover2025} performs
context-preserving verification via KV-cache override, reducing the flip-flop
behaviour where previously committed tokens are repeatedly revised. A complementary
direction trains a lightweight auxiliary model to predict which token positions
have converged, skipping denoising steps for stable positions and concentrating
computation where it is still needed~\citep{ye2025dream}. SAID builds on this
family of ideas by combining a structured two-stage unmasking schedule with
confidence-threshold stratification, achieving acceleration without any auxiliary
model, revocation mechanism, or additional training.

%-----------------------------------------------------------------------
\section{Methodology}
\label{sec:method}
%-----------------------------------------------------------------------

\begin{figure*}[t]
\vspace{-15pt}
  \centering
  \begin{subfigure}[b]{\linewidth}
    \includegraphics[width=\textwidth]{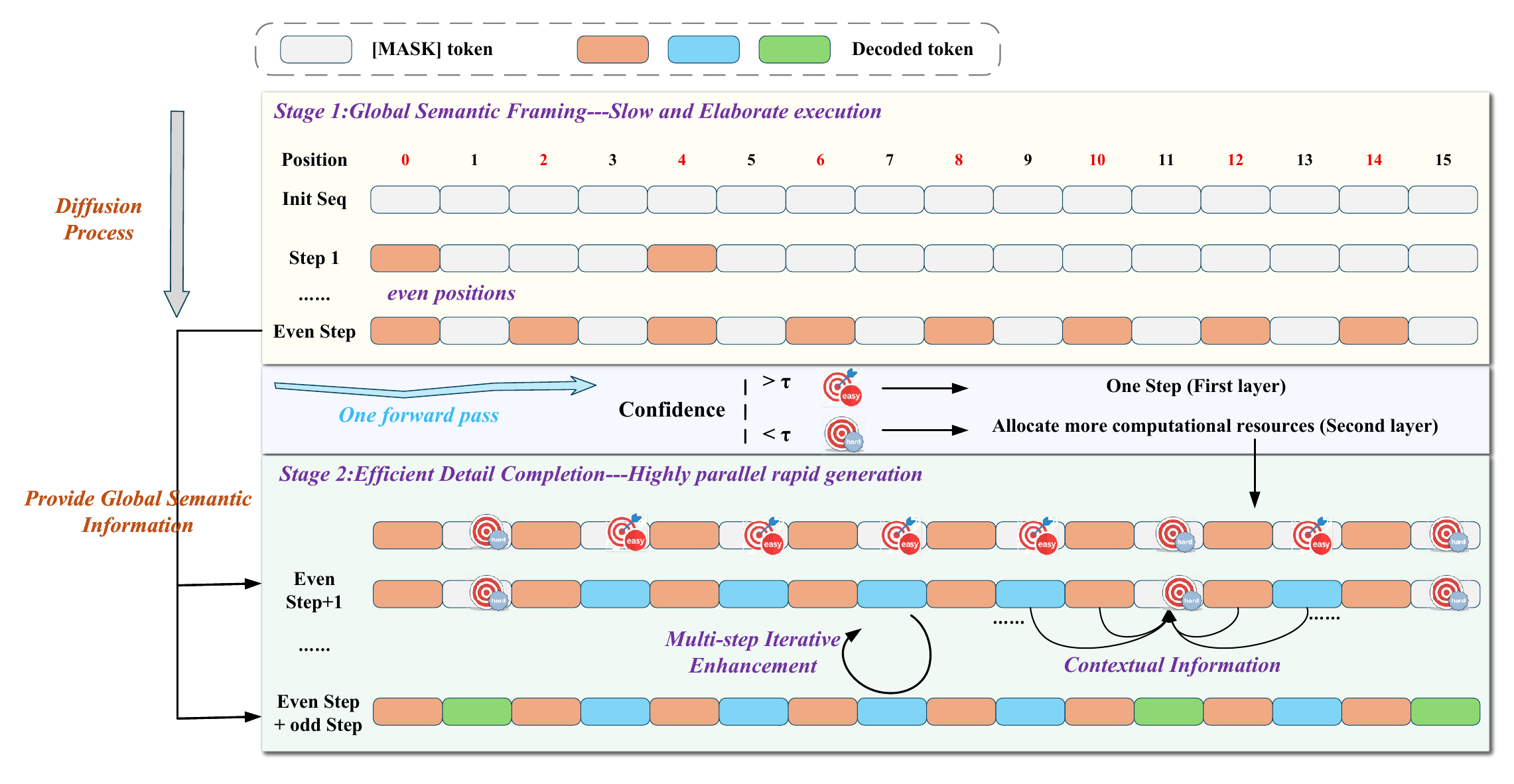}
    \caption{Two-stage decoding: even positions undergo $T$-step semantic
      modeling; odd positions are filled in fewer step via CHLG.}
    \label{fig:said_method}
  \end{subfigure}

  \vspace{6pt}

  \begin{subfigure}[b]{\linewidth}
    \includegraphics[width=\textwidth]{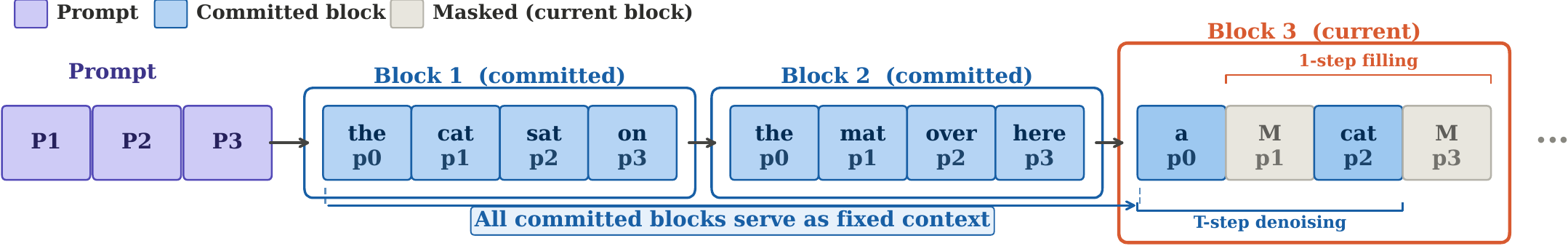}
    \caption{Block-level inference: blocks are processed left to right;
      committed blocks serve as fixed context for the current block.}
    \label{fig:said_block}
    \vspace{-5pt}
  \end{subfigure}
\vspace{-0.4cm}
  \caption{Overview of the Scaffold-Aware Iterative Decoding (SAID) Framework
for Accelerating Diffusion Language Model Inference.}
  \label{fig:said_overview}
\end{figure*}

\subsection{Motivation and Overview}

Autoregressive models generate text sequentially, accumulating context one token
at a time but incurring high decoding latency. Standard masked diffusion models
reduce this sequential bottleneck by predicting all positions in parallel, but they
typically allocate the same denoising budget to every token. This uniform allocation
is inefficient: structurally important tokens may require multiple refinement steps
to establish global coherence, whereas tokens that are strongly constrained by their
surrounding context can often be resolved with far fewer updates. We build on the masked diffusion framework reviewed in
Appendix~\ref{sec:preliminary} and propose SAID, a scaffold-aware
two-stage decoding schedule that reallocates denoising computation
across token positions.

Figure~\ref{fig:said_method} and Figure~\ref{fig:said_block} illustrates the overall SAID framework. Figure~\ref{fig:said_method} shows token-level SAID, where the model first performs iterative denoising on uniformly interleaved anchor tokens to construct a semantic scaffold, and then completes the remaining detail tokens with fewer denoising steps. Figure~\ref{fig:said_block} shows block-level SAID, where the same anchor-detail schedule is applied within each block and previously committed blocks serve as fixed left context. Finally, Confidence-Hierarchical Layered Generation separates detail tokens by confidence, committing high-confidence tokens immediately and allocating additional denoising steps only to uncertain tokens. Together, these components reduce unnecessary denoising computation while preserving generation quality.
\subsection{Token-Level SAID}

To operationalize the coarse-to-fine principle at the token level, SAID uses an anchor-guided two-stage decoding schedule consisting of a semantic modeling stage and a detail filling stage, as shown in Figure~\ref{fig:said_method}. The \textbf{semantic modeling stage} concentrates the full iterative denoising budget on a subset of anchor positions, while the \textbf{detail filling stage} resolves the remaining positions in a single forward pass conditioned on the committed anchors. By shifting iterative computation from all tokens to anchors, SAID reduces decoding latency without requiring additional training.

\noindent\textbf{Semantic Modeling Stage.}
Let $\mathbf{x} = (x_1, x_2, \ldots, x_n)$ denote the generation sequence, excluding the prompt. The semantic modeling stage partitions the positions into two disjoint parity-based subsets:
\begin{equation}
  \mathcal{E} = \{i : i \bmod 2 = 0\}, \quad
  \mathcal{O} = \{i : i \bmod 2 = 1\}.
\end{equation}
The even-indexed set $\mathcal{E}$ is used as the anchor set. SAID applies the standard masked-diffusion denoising loop only to these anchor positions. Given a budget of $T$ denoising steps, each step $t \in \{1,\ldots,T\}$ runs a forward pass and computes a confidence score for every currently masked anchor position:
\begin{equation}
  c_j = \max_v p_\theta(v \mid \mathbf{x}), \quad
  j \in \mathcal{E}_{\mathrm{masked}}.
\end{equation}
The model then commits the highest-confidence positions according to
\begin{equation}
  n_t^{\mathcal{E}} =
  \left\lfloor
  \frac{|\mathcal{E}_{\mathrm{masked}}|}{T - t + 1}
  \right\rfloor,
\end{equation}
and fixes their predicted values for subsequent steps. This confidence-based unmasking follows the inference procedure used in standard LLaDA decoding. After $T$ steps, all positions in $\mathcal{E}$ are committed, forming an interleaved anchor scaffold for the remaining tokens.

\noindent\textbf{Detail Filling Stage.}
After all anchor positions in $\mathcal{E}$ are committed, the detail filling stage predicts every remaining position in $\mathcal{O}$ with a single forward pass:
\begin{equation}
  \hat{x}_j = \arg\max_v p_\theta(v \mid \mathbf{x}_{\mathcal{E}}),
  \quad \forall j \in \mathcal{O}.
\end{equation}
Conditioned on the fixed anchor context, this one-step filling pass replaces iterative denoising over the remaining tokens. As a result, SAID shifts computation from uniform denoising over all positions to staged denoising over anchors followed by lightweight detail completion. This reduces the effective denoising workload without additional training. We study the resulting quality--latency trade-off in Section~\ref{sec:experiments}.

\subsection{Confidence-Hierarchical Layered Generation}
\label{sec:confidence}

The one-step detail filling stage in SAID treats all detail tokens uniformly, even though their prediction difficulty can vary substantially. Some odd-position tokens are strongly constrained by the committed anchor context and can be predicted with high confidence, while others remain ambiguous and require additional refinement. Committing all detail tokens in a single pass may therefore introduce errors on low-confidence positions and degrade generation quality. To address this issue, we introduce a confidence-hierarchical layered generation mechanism that organizes detail tokens into confidence-based layers: high-confidence tokens are committed immediately, while low-confidence tokens are deferred to a refinement layer.

Given the committed anchor positions $\mathcal{E}$, we first run the same forward pass used by the base detail filling stage. For each odd position $j \in \mathcal{O}$, we compute its peak confidence
\begin{equation}
  c_j = \max_v p_\theta(v \mid \mathbf{x}_{\mathcal{E}}).
\end{equation}
A confidence threshold $\lambda \in (0,1)$ controls how conservatively the model commits odd-position tokens: higher values defer more tokens to the refinement layer, while lower values make the procedure closer to base SAID. We use this threshold to partition the odd positions into easy and hard subsets:
\begin{equation}
  \mathcal{O}_{\mathrm{easy}} = \{j \in \mathcal{O} : c_j \geq \lambda\}, \quad
  \mathcal{O}_{\mathrm{hard}} = \mathcal{O} \setminus \mathcal{O}_{\mathrm{easy}}.
\end{equation}
Tokens in $\mathcal{O}_{\mathrm{easy}}$ form the high-confidence layer and are committed immediately by a forward pass. Since this pass is already required by base SAID, constructing the confidence hierarchy itself introduces no additional model evaluation.

The remaining tokens in $\mathcal{O}_{\mathrm{hard}}$ form the low-confidence refinement layer. They are refined with an additional denoising budget of $T_{\mathrm{hard}}$ steps, conditioned on both the committed anchors and the newly committed high-confidence detail tokens:
\begin{equation}
\begin{aligned}
  \hat{x}_j^{(t+1)}
  &= \mathrm{Denoise}\!\left(
    \mathbf{x}_{\mathcal{E}} \cup \mathbf{x}_{\mathcal{O}_{\mathrm{easy}}},
    \mathcal{O}_{\mathrm{hard}}, t
  \right), \\
  &\quad j \in \mathcal{O}_{\mathrm{hard}}, \quad
  t = 1,\ldots,T_{\mathrm{hard}}.
\end{aligned}
\end{equation}
This layered refinement allocates extra computation only to uncertain detail tokens, while preserving one-step filling for high-confidence tokens. When most detail tokens fall into the high-confidence layer, the additional cost remains small while still allowing the model to refine low-confidence predictions.

\subsection{Block-Level SAID}
\label{sec:block_said}

Blockwise diffusion is widely used in diffusion language model inference because it offers a strong balance between generation quality and decoding efficiency. Motivated by this setting, we extend SAID from token-level inference to block-level inference. The generation sequence is divided into consecutive blocks, and the blocks are processed from left to right. Within \textbf{each block}, SAID applies the same anchor-guided two-stage schedule, while later blocks can condition on the fully committed tokens from earlier blocks.

Specifically, we process blocks from left to right and apply SAID independently within \textbf{each block}. For block $\mathcal{B}_b$, all preceding blocks have already been committed and serve as fixed left context. The positions in $\mathcal{B}_b$ are partitioned into an anchor set $\mathcal{E}^{(b)}$ and a detail set $\mathcal{O}^{(b)}$. Instead of using the full blockwise denoising budget on all positions, Block-SAID allocates a reduced anchor-denoising budget $T_b^{\mathcal{E}}$ to $\mathcal{E}^{(b)}$ and commits anchor tokens in confidence-ranked order. The detail tokens in $\mathcal{O}^{(b)}$ are then resolved using the confidence-guided refinement procedure in Section~\ref{sec:confidence}: high-confidence detail tokens are committed after one stratification pass, while low-confidence detail tokens receive $T_{\mathrm{hard}}$ additional refinement steps.

Under the standard blockwise diffusion schedule, each block requires $T_b$ denoising steps to progressively resolve all positions in $\mathcal{B}_b$, giving a total cost of
\begin{equation}
  C_{\mathrm{Block}} = B \cdot T_b.
\end{equation}
In contrast, Block-SAID uses $T_b^{\mathcal{E}}$ anchor-denoising steps, one detail-filling pass, and $T_{\mathrm{hard}}$ optional hard-token refinement steps per block:
\begin{equation}
  C_{\mathrm{Block\text{-}SAID}} = B \cdot \left(T_b^{\mathcal{E}} + 1 + T_{\mathrm{hard}}\right).
\end{equation}
When $T_b^{\mathcal{E}} + 1 + T_{\mathrm{hard}} < T_b$, Block-SAID reduces the number of decoding steps relative to standard blockwise diffusion. In practice, setting $T_b^{\mathcal{E}} \approx T_b/2$ yields a per-block schedule of approximately $T_b/2 + 1 + T_{\mathrm{hard}}$ steps, because only anchor positions are refined iteratively while most detail tokens are filled in one pass.

\section{Experiments}
\label{sec:experiments}
%-----------------------------------------------------------------------

\subsection{Experimental Setups}
\noindent\textbf{Implementation Details.}
Full details of hardware configuration, hyperparameter settings,
and evaluation protocols are provided in Appendix~\ref{sec:impl}.

% \paragraph{Models.}
% We evaluate SAID on two representative masked diffusion language models:
% LLaDA-8B~\citep{nie2025llada}, a standard token-level masked diffusion model,
% and LLaDA~1.5, a block-diffusion variant that processes the generation sequence
% in fixed-length blocks. Both models are used without any fine-tuning or
% modification to model weights, confirming that SAID is a purely inference-time,
% training-free acceleration strategy.

% \paragraph{Benchmarks.}
% We evaluate SAID on six benchmarks covering code generation, scientific reasoning, commonsense reasoning, mathematical reasoning, and multi-domain knowledge. Specifically, we use MBPP~\citep{austin2021mbpp} for code generation, GPQA~\citep{rein2023gpqa} for graduate-level science reasoning, ARC-C~\citep{clark2018arc} for commonsense reasoning, GSM8K~\citep{cobbe2021gsm8k} and MATH~\citep{hendrycks2021math} for mathematical reasoning, and MMLU-Pro~\citep{wang2024mmlupro} for multi-domain knowledge evaluation. We report pass@1 for MBPP and accuracy for all other benchmarks.

% \paragraph{Baselines and Metrics.}
% We compare SAID against the standard LLaDA decoding baseline using a uniform
% step budget of $T = 100$ denoising steps. For block-level experiments, we
% compare SAID against the LLaDA~1.5 baseline under varying block lengths
% $\ell \in \{16, 32, 64\}$. We report task accuracy, throughput (tokens/s),
% and per-sample latency (s). Speedup is computed as the ratio of baseline
% latency to SAID latency.

\subsection{Token-Level Results}

\begin{figure}[t]
  \centering
  \includegraphics[width=\linewidth]{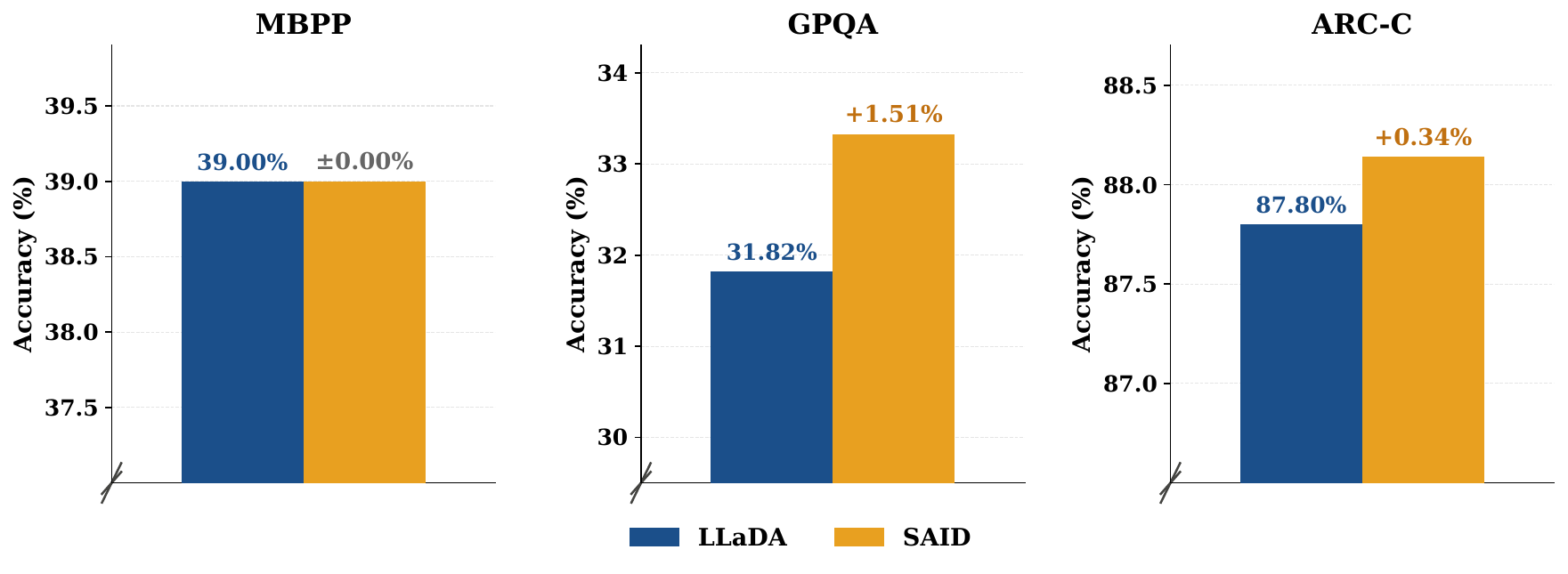}
  \caption{Accuracy comparison of LLaDA and SAID across benchmarks.
    Percentages above SAID bars indicate the delta relative to LLaDA.}
  \label{fig:accuracy}
\end{figure}

\begin{figure}[t]
  \centering
  \includegraphics[width=\linewidth]{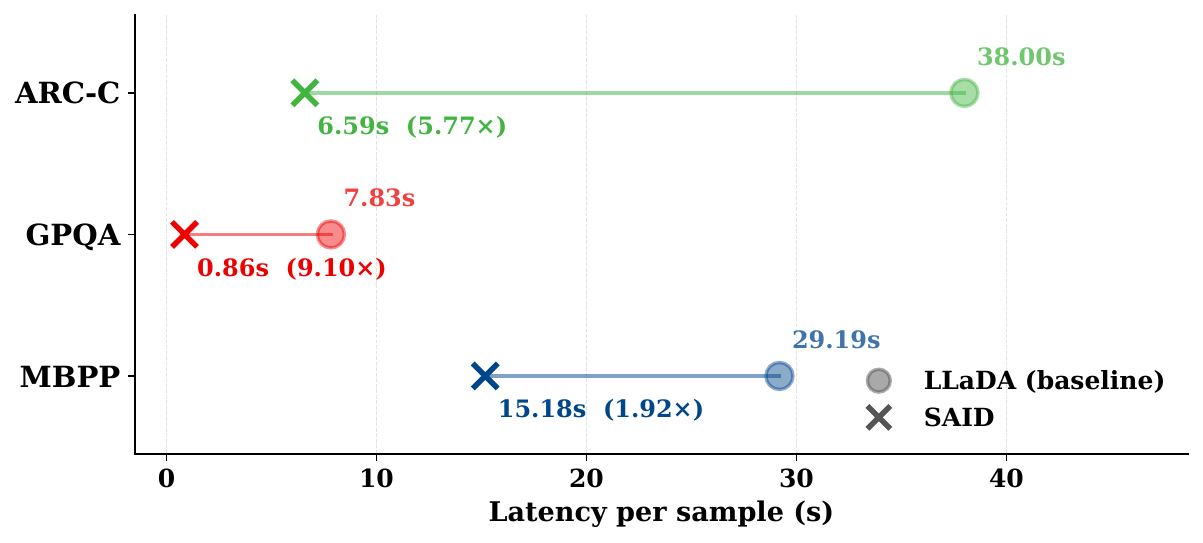}
  \caption{Latency comparison of LLaDA and SAID and per-benchmark
    speedup ratio }
  \label{fig:latency1}
  \vspace{-12pt}
\end{figure}

\begin{figure}[t]
  \centering
  \includegraphics[width=\linewidth]{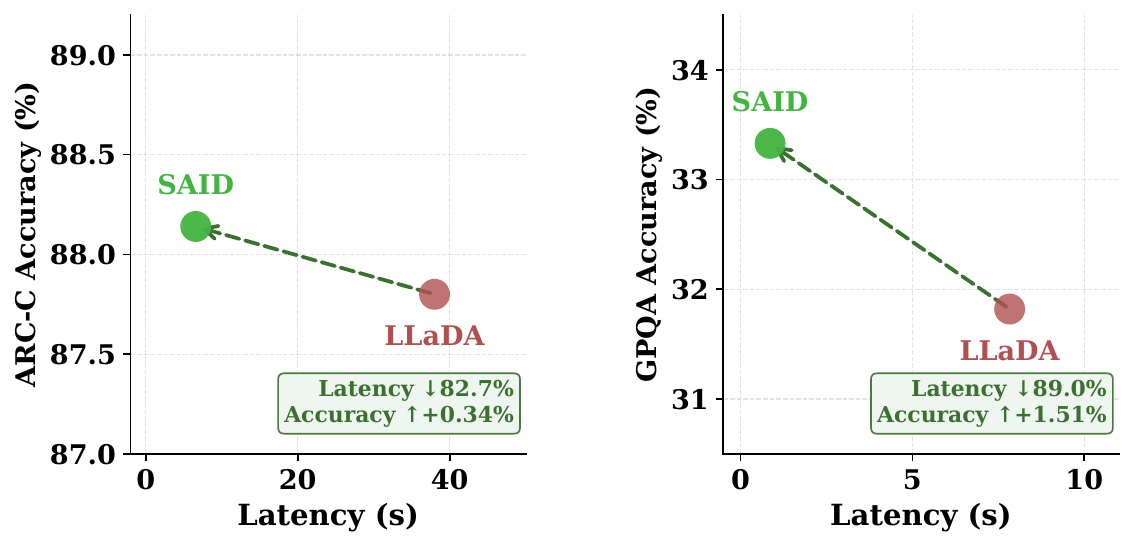}
  \caption{Accuracy--latency trade-off of LLaDA and SAID on ARC-C (left)
    and GPQA (right). Each point represents a method; arrows indicate the
    direction of improvement from LLaDA to SAID. }
  \label{fig:latency2}
\end{figure}

Table~\ref{tab:performance} reports the accuracy, throughput, and latency of SAID compared with standard LLaDA decoding. SAID consistently improves inference efficiency across all benchmarks while maintaining competitive task performance. The largest gains are observed on GPQA and ARC-C, where SAID achieves $9.10\times$ and $5.77\times$ latency reduction, respectively. On longer-generation tasks such as GSM8K and MBPP, the speedups are more moderate, reaching $1.97\times$ and $1.92\times$, but still provide substantial acceleration. Notably, SAID also slightly improves accuracy on GPQA and ARC-C by $+1.51\%$ and $+0.34\%$, respectively, suggesting that scaffold-aware decoding can preserve, and in some cases even improve, the quality of structured reasoning. Figure~\ref{fig:accuracy}, Figure~\ref{fig:latency1} and Figure~\ref{fig:latency2} further visualize the accuracy and latency trends.

\begin{table*}[tp]
\centering
\caption{Performance Comparison of LLaDA and SAID(with CHLG)}
\label{tab:performance}
\vspace{-5pt}
\setlength{\tabcolsep}{10pt}
\begin{tabular}{llccc}
\toprule
Benchmark & Method & Accuracy & Throughput (tokens/s) & Latency  \\
\midrule
MBPP     & LLaDA & 39.00\% & 8.78  & 29.19  \\
         & SAID  & 39.00\% (+0.00\%) & 16.88 \textbf{(1.92$\times$)} & 15.18 \textbf{(1.92$\times$)} \\
GPQA     & LLaDA & 31.82\% & 8.18  & 7.83   \\
         & SAID  & 33.33\% (+1.51\%) & 74.50 \textbf{(9.11$\times$)} & 0.86  \textbf{(9.10$\times$)} \\
ARC-C    & LLaDA & 87.80\% & 13.48 & 38.00  \\
         & SAID  & 88.14\% (+0.34\%) & 77.71 \textbf{(5.76$\times$)} & 6.59  \textbf{(5.77$\times$)} \\
GSM8K    & LLaDA & 68.99\% & 4.46  & 114.85 \\
         & SAID  & 68.23\% (-0.76\%) & 8.78  \textbf{(1.97$\times$)} & 58.33 \textbf{(1.97$\times$)} \\
Math     & LLaDA & 29.94\% & 13.10 & 39.13  \\
         & SAID  & 29.48\% (-0.46\%) & 25.74 \textbf{(1.96$\times$)} & 19.95 \textbf{(1.96$\times$)} \\
MMLU-pro & LLaDA & 36.70\% & 11.34 & 22.60  \\
         & SAID  & 35.60\% (-1.10\%) & 19.48 \textbf{(1.72$\times$)} & 13.15 \textbf{(1.72$\times$)} \\
\bottomrule
\end{tabular}
\vspace{-20pt}
\end{table*}

\begin{table*}[tp]
\centering
\caption{Performance comparison of LLaDA 1.5 and SAID across benchmarks.
  Speedup (in parentheses) is computed relative to LLaDA 1.5.}
    \vspace{-5pt}
\label{tab:block}
\setlength{\tabcolsep}{6pt}
\renewcommand{\arraystretch}{0.92}
\begin{tabular}{llcclll}
\toprule
\textbf{Benchmark} & \textbf{Gen.} & \textbf{Block} & \textbf{Method}
  & \textbf{Accuracy}
  & \textbf{Throughput (tok/s)}
  & \textbf{Latency (s)} \\
\midrule
\multirow{6}{*}{ARC-C}
  & \multirow{6}{*}{256} & \multirow{2}{*}{16}
  & LLaDA 1.5 & 86.44\% & 5.45  & 46.97  \\
  & & & SAID      & 88.14\% & 10.13 \textbf{(1.86$\times$)} & 25.00 \textbf{(1.88$\times$)} \\
  \cmidrule{3-7}
  & & \multirow{2}{*}{32}
  & LLaDA 1.5 & 86.10\% & 6.01  & 42.59  \\
  & & & SAID      & 85.42\% & 14.34 \textbf{(2.38$\times$)} & 17.86 \textbf{(2.38$\times$)} \\
  \cmidrule{3-7}
  & & \multirow{2}{*}{64}
  & LLaDA 1.5 & 86.10\% & 7.44  & 34.39  \\
  & & & SAID      & 85.76\% & 17.13 \textbf{(2.30$\times$)} & 14.95 \textbf{(2.30$\times$)} \\
\midrule
\multirow{6}{*}{GPQA}
  & \multirow{6}{*}{256} & \multirow{2}{*}{16}
  & LLaDA 1.5 & 33.84\% & 3.43  & 74.67  \\
  & & & SAID      & 33.33\% & 6.18 \textbf{(1.80$\times$)}  & 41.43 \textbf{(1.80$\times$)} \\
  \cmidrule{3-7}
  & & \multirow{2}{*}{32}
  & LLaDA 1.5 & 32.83\% & 3.16  & 81.03  \\
  & & & SAID      & 31.82\% & 4.67 \textbf{(1.48$\times$)}  & 54.82 \textbf{(1.48$\times$)} \\
  \cmidrule{3-7}
  & & \multirow{2}{*}{64}
  & LLaDA 1.5 & 32.83\% & 2.55  & 100.21 \\
  & & & SAID      & 30.81\% & 4.39 \textbf{(1.72$\times$)}  & 58.34 \textbf{(1.72$\times$)} \\
\midrule
\multirow{2}{*}{GSM8K}
  & \multirow{2}{*}{256} & \multirow{2}{*}{16}
  & LLaDA 1.5 & 82.79\% & 2.50  & 102.49 \\
  & & & SAID      & 80.36\% & 7.02 \textbf{(2.81$\times$)}  & 36.47 \textbf{(2.81$\times$)} \\
\bottomrule
\end{tabular}
\end{table*}

\begin{table*}[tp]
\centering
\caption{Ablation Study of SAID and CHLG Modules}
\label{tab:ablation}
\vspace{-5pt}
\setlength{\tabcolsep}{16pt}
\begin{tabular}{cccccc}
\toprule
SAID & CHLG & MBPP & GPQA & ARC-C & GSM8K \\
\midrule
$\times$ & $\times$ & 39.00\% & 31.82\% & 87.80\% & \textbf{68.99\%} \\
\checkmark & $\times$ & 36.40\% & 32.32\% & 87.46\% & 68.08\% \\
\checkmark & \checkmark & \textbf{39.00\%} & \textbf{33.33}\% & \textbf{88.14\%} & 68.23\% \\
\bottomrule
\end{tabular}
\end{table*}

\begin{table}[t]
  \centering
  \caption{Comparison of Different Thresholds in CHLG}
  \label{tab:threshold}
  \begin{tabular}{ccccc}
    \toprule
    $\lambda$ & ARC-C & MBPP & GPQA & GSM8K \\
    \midrule
    0.6 & 87.80\% & 38.60\% & 33.33\% & 68.16\% \\
    0.7 & 88.14\% & \textbf{38.80\%} & 33.33\% & 68.08\% \\
    0.8 & 88.14\% & 38.60\% & 33.33\% & \textbf{68.23\%} \\
    0.9 & 88.14\% & 38.20\% & 33.33\% & 68.08\% \\
    \bottomrule
  \end{tabular}
    \vspace{-10pt}
\end{table}

\begin{table}[t]
\centering
\caption{Performance Comparison of Different Decoding Methods}
\label{tab:selection}
\setlength{\tabcolsep}{6pt}
\begin{tabular}{lccc}
\toprule
 & SAID & SAID* & Random \\
\midrule
MBPP & \textbf{36.40\%} & 34.60\% & 34.20\% \\
ARC-C & 87.46\% & \textbf{89.15\%} & 87.23\% \\
GPQA & \textbf{32.32\%} & 27.78\% & 31.14\% \\
\bottomrule
\end{tabular}
  \vspace{-10pt}
\end{table}

\subsection{Block-Level Results}
Table~\ref{tab:block} reports results for Block-SAID applied to LLaDA~1.5 under
varying block lengths. On ARC-C, Block-SAID with block length 16 achieves
$1.88\times$ latency speedup while improving accuracy from 86.44\% to 88.14\%.
As block length increases to 32 and 64, the speedup grows to $2.38\times$ and
$2.30\times$ respectively, though with a slight decrease in accuracy, reflecting
the standard quality--efficiency trade-off in block diffusion. On GPQA, Block-SAID
consistently accelerates inference across all block lengths, with speedups ranging
from $1.48\times$ to $1.80\times$. The largest absolute speedup is observed on
GSM8K, where Block-SAID reduces per-sample latency from 102.49s to 36.47s
($2.81\times$).SAID accelerates inference by rapidly generating approximately 50\% of tokens in the second stage within each block with minimal reconstruction steps. However, GSM8K is a mathematical reasoning task that heavily relies on long‑range logical reasoning and step‑wise coherence, which consequently leads to accuracy degradation. Nevertheless, our model achieves a superior speed‑accuracy trade‑off. 

\subsection{Ablation Studies}
\noindent\textbf{Contribution of SAID and CHLG.}
Table~\ref{tab:ablation} reports the ablation of each component. Using SAID
alone (without CHLG) already improves throughput significantly, but slightly
reduces accuracy on MBPP ($-2.60\%$) and GSM8K ($-0.91\%$) relative to
the baseline, suggesting that committing all odd-position tokens in a single
step can occasionally introduce errors on structurally complex positions.
Adding CHLG fully recovers the accuracy drop on MBPP (back to 39.00\%) and
further improves GPQA by $+1.51\%$ and ARC-C by $+0.34\%$, confirming that
the confidence-hierarchical refinement mechanism effectively compensates for
the one-step detail-filling approximation.

\begin{figure}[t]
  \centering
  \includegraphics[width=\linewidth]{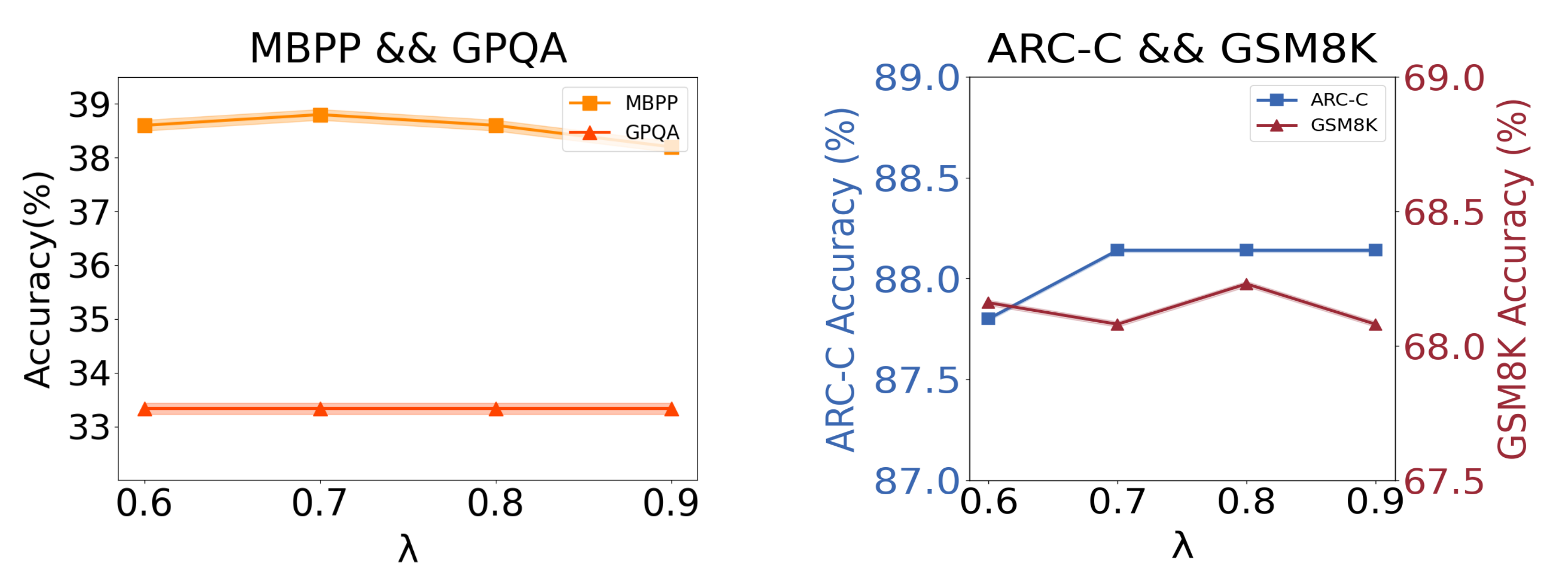}
  \caption{Comparison of Different Thresholds in CHLG.}
  \label{fig:Threshold}
\end{figure}

\noindent\textbf{Effect of Confidence Threshold $\lambda$.}
Table~\ref{tab:threshold} reports accuracy across four benchmarks as the CHLG threshold $\lambda$
varies from 0.6 to 0.9. Results are stable across this range, with ARC-C,
GPQA, and GSM8K showing negligible variation. MBPP shows mild sensitivity,
peaking at $\lambda = 0.7$ (38.80\%).
Figure~\ref{fig:Threshold} visualizes these trends.

\begin{figure}[t!]
  \centering
  \includegraphics[width=\linewidth]{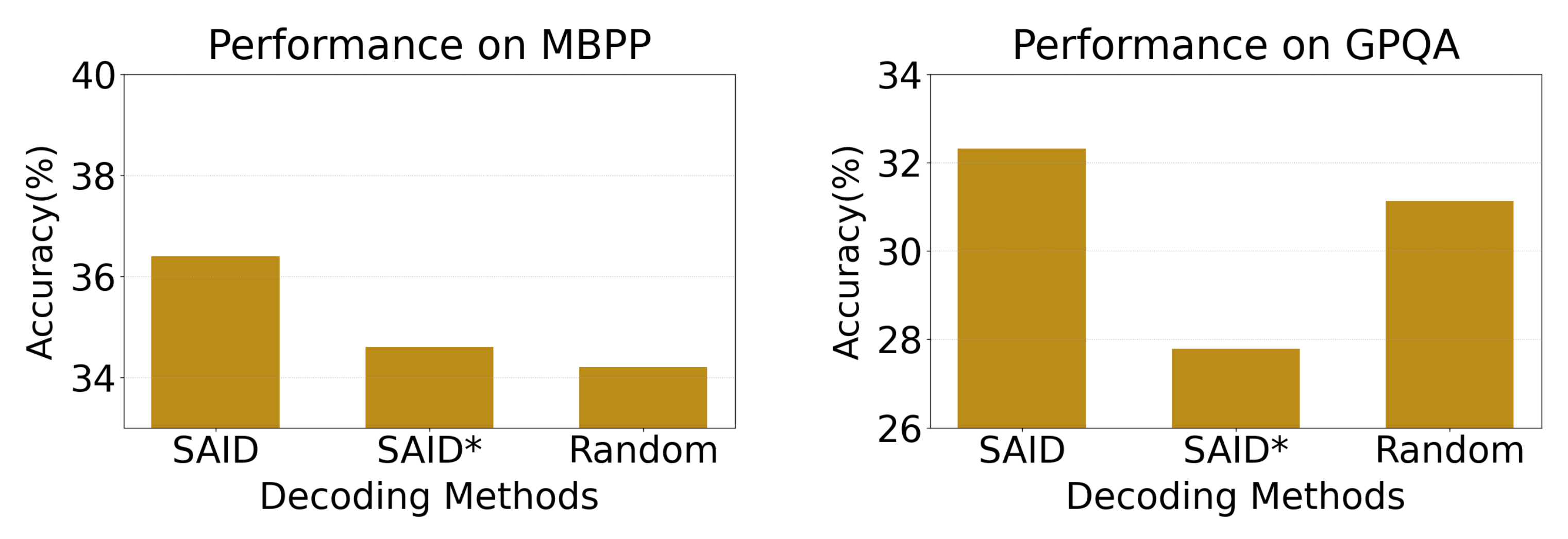}

  \caption{Performance Comparison of Different Decoding Methods.}
  \label{fig:decoding}
\end{figure}

\noindent\textbf{Decoding Strategy Comparison.}
Table~\ref{tab:selection} compares SAID against alternative decoding strategies: SAID* (a variant
with reversed parity, filling odd positions first), and Random (first decode half of the tokens randomly, then generate the other half in one step). SAID achieves the best overall balance across benchmarks.
SAID* performs well on ARC-C (89.15\%) but poorly on GPQA (27.78\%),
suggesting that the choice of which positions serve as anchors matters for
reasoning-intensive tasks. 
Figure~\ref{fig:decoding} visualizes the comparison.

%------------------------------------------------

\section{Conclusion}
%-----------------------------------------------------------------------

We present SAID, a training-free acceleration framework for masked diffusion
language models that decomposes inference into a semantic modeling stage and a single-pass detail filling stage. The proposed Confidence-Hierarchical Layered Generation
(CHLG) mechanism further refines low-confidence detail tokens with additional
denoising steps, improving accuracy on challenging positions without incurring
significant overhead. Extended to block-wise diffusion decoding, Block-SAID
applies the same two-stage schedule within each block while leveraging
committed blocks as fixed context, reducing total forward passes across
diverse benchmarks. Experiments on LLaDA-8B and LLaDA~1.5 demonstrate
consistent speedups of up to $9.1\times$ with competitive task performance.
%-----------------------------------------------------------------------
\section{Limitations}
%-----------------------------------------------------------------------

\noindent\textbf{Data and Language Coverage.}
Our experiments are conducted exclusively on English-language benchmarks.
The effectiveness of SAID on non-English or multilingual generation tasks
remains unverified, as the even/odd scaffold structure may interact
differently with morphologically richer or character-based languages.
Additionally, the benchmarks used---MBPP, GPQA, ARC-C, GSM8K, MATH,
and MMLU-Pro---cover coding, mathematics, and commonsense domains
but do not represent open-ended generation, dialogue, or low-resource
settings where masked diffusion models may behave differently.

\noindent\textbf{Evaluation Metric Blind Spots.}
All benchmarks are evaluated using accuracy or pass@1, which capture
correctness but not fluency, coherence, or calibration. SAID's
scaffold-aware decoding may affect the distribution of generated text
in ways that are not reflected by these metrics, particularly for
open-ended tasks where generation quality is multidimensional.

\noindent\textbf{Speedup Bounded by Fixed Partition.}
The speedup of SAID is inherently bounded by the fixed 50\% even/odd
partition. On longer-generation tasks such as MBPP (512 tokens), the
gain is more modest ($1.92\times$) compared to short-answer benchmarks
such as GPQA ($9.1\times$), because the absolute number of saved
denoising steps is diluted over a larger total budget. Adaptive
scaffold selection that dynamically adjusts the proportion of anchor
positions remains an open problem.

%-----------------------------------------------------------------------
% \bibliographystyle{unsrtnat}
% \bibliography{custom}
%-----------------------------------------------------------------------

% \bibliographystyle{unsrtnat} 
% \bibliographystyle{apalike} 
\bibliography{custom}

\clearpage
\newpage
\appendix

%-----------------------------------------------------------------------
\section{Preliminary}
\label{sec:preliminary}
%-----------------------------------------------------------------------

\subsection{Diffusion Language Modeling}

Let $\mathbf{x}^0 = (x_1^0, \ldots, x_n^0) \in \mathcal{V}^n$ denote a sequence
of $n$ discrete tokens. A diffusion language model
(DLM)~\citep{austin2021d3pm,sahoo2024mdlm,lou2024sedd} defines a forward
noising process $q(\mathbf{x}^t \mid \mathbf{x}^0)$ indexed by a noise level
$t \in [0,1]$. In masked diffusion, each token is independently replaced by a
special mask token \texttt{[M]}:
\begin{equation}
  q(x_i^t \mid x_i^0) =
  \begin{cases}
    1 - \alpha_t, & \text{if } x_i^t = x_i^0, \\
    \alpha_t,     & \text{if } x_i^t = \texttt{[M]},
  \end{cases}
\end{equation}
where $\alpha_t$ is a monotonically increasing masking rate with
$\alpha_0 = 0$ and $\alpha_1 = 1$. The joint corruption distribution factorizes
as $q(\mathbf{x}^t \mid \mathbf{x}^0)=\prod_{i=1}^n q(x_i^t \mid x_i^0)$.

The reverse process is modeled by a bidirectional Transformer
$p_\theta(\mathbf{x}^0 \mid \mathbf{x}^t)$, which predicts the original tokens
at masked positions in parallel. Given
$\mathcal{M}^t = \{i : x_i^t = \texttt{[M]}\}$, the per-sample masked prediction
loss is
\begin{equation}
  \ell_{\mathrm{DLM}}(\theta; \mathbf{x}^0, \mathbf{x}^t)
  =
  - \frac{1}{|\mathcal{M}^t|}
  \sum_{i \in \mathcal{M}^t}
  \log p_\theta(x_i^0 \mid \mathbf{x}^t),
\end{equation}
and the training objective is
\begin{equation}
  \mathcal{L}_{\mathrm{DLM}}(\theta)
  =
  \mathbb{E}_{t,\,\mathbf{x}^t}
  \left[
    \ell_{\mathrm{DLM}}(\theta; \mathbf{x}^0, \mathbf{x}^t)
  \right],
\end{equation}
where $t \sim \mathcal{U}[0,1]$ and
$\mathbf{x}^t \sim q(\cdot \mid \mathbf{x}^0)$.

At inference, generation starts from the fully masked sequence
$\mathbf{x}^1 = \texttt{[M]}^n$ and proceeds over $T$ denoising steps
$t_T > \cdots > t_0 = 0$. At each step, the model predicts token distributions
for all masked positions, and a subset
$\mathcal{U}_k \subseteq \mathcal{M}^{t_k}$ is selected for unmasking. The choice of $\mathcal{U}_k$, referred to as the \emph{unmasking schedule},
governs the quality--efficiency trade-off in iterative generation. SAID builds
on this observation by selecting which tokens to unmask at each
denoising step.

\subsection{Block Diffusion}

Block diffusion~\citep{arriola2025bd3} decomposes the generation sequence into $B$
consecutive blocks of equal length $\ell$, processed in a fixed left-to-right order.
Let the prompt be $\mathbf{p}\in\mathcal{V}^P$ and the generation sequence be
$\mathbf{x}=[\mathbf{x}^{(1)},\ldots,\mathbf{x}^{(B)}]\in\mathcal{V}^{B\ell}$.
The joint distribution is factorized causally across blocks:
\begin{equation}
  p_\theta(\mathbf{x}\mid\mathbf{p})
  = \prod_{b=1}^{B} p_\theta\!\left(\mathbf{x}^{(b)} \mid
    \mathbf{p},\, \mathbf{x}^{(1)},\ldots,\mathbf{x}^{(b-1)}\right).
\end{equation}
Each factor $p_\theta(\mathbf{x}^{(b)} \mid \mathbf{p}, \mathbf{x}^{(<b)})$ is an
independent masked diffusion process over the $\ell$ positions in block $b$,
conditioned on the prompt and all preceding committed blocks as fixed context.
Generation proceeds block by block: block $b$ runs its full $T_b$ denoising steps
to completion before block $b+1$ begins, ensuring that each block benefits from a
fully resolved left context. Within each block, attention remains bidirectional over
all $\ell$ positions, preserving the global coherence of the underlying diffusion
model. SAID inherits this block-sequential structure and further reduces the number
of denoising steps $T_b$ required per block by applying the two-stage
strategy within each block independently.

\section{Implementation Details}
\label{sec:impl}

\subsection{Software}

The base models LLaDA-8B and LLaDA~1.5 are loaded from their official
HuggingFace checkpoints without any modification to model weights or
architecture.

\subsection{Block-Level Configuration}

For block-level experiments with LLaDA~1.5, the generation length is set to
256 tokens for ARC-C, GPQA, and GSM8K.
Block lengths $\ell \in \{16, 32, 64\}$ are evaluated independently.
Within each block, SAID applies the same even/odd two-stage schedule with
$T_b = T / B$ steps allocated to anchor positions, where $B = L / \ell$
is the number of blocks and $L$ is the generation length.
The CHLG threshold $\lambda = 0.8$ and $T_{\mathrm{hard}} = 5$ are used
for block-level experiments.

\subsection{Benchmark Evaluation}

We evaluate SAID on six benchmarks covering code generation, scientific
reasoning, commonsense reasoning, mathematical reasoning, and multi-domain
knowledge. Specifically, we use MBPP~\citep{austin2021mbpp} for code
generation, GPQA~\citep{rein2023gpqa} for graduate-level science reasoning,
ARC-C~\citep{clark2018arc} for commonsense reasoning,
GSM8K~\citep{cobbe2021gsm8k} and MATH~\citep{hendrycks2021math} for
mathematical reasoning, and MMLU-Pro~\citep{wang2024mmlupro} for
multi-domain knowledge evaluation. We report pass@1 for MBPP and accuracy
for all other benchmarks.

\subsection{Latency Measurement}

Latency is measured as wall-clock time per sample, averaged over the
full benchmark test set. We use \texttt{torch.cuda.synchronize()} before
and after each generation call to ensure accurate GPU timing.
Throughput (tokens/s) is computed as generated tokens divided by
total wall-clock time across the full benchmark.
All measurements exclude tokenization and prompt encoding time.

\end{document}